\pdfoutput=1

\documentclass[11pt]{article}

\usepackage[]{ACL2023}

\usepackage{times}
\usepackage{latexsym}

\usepackage[T1]{fontenc}

\usepackage[utf8]{inputenc}

\usepackage{microtype}

\usepackage{inconsolata}
\usepackage{subfigure}
\usepackage{CJKutf8}
\usepackage{graphicx}
\usepackage[normalem]{ulem}
\usepackage{soul}
\usepackage{color}
\usepackage{xcolor}
\usepackage{multirow}
\usepackage{makecell}
\usepackage{xspace}  
\usepackage{amssymb}
\usepackage{amsmath}
\usepackage{booktabs}
\usepackage[symbol]{footmisc}

\definecolor{lightgreen}{HTML}{b7ea67}
\definecolor{goodyellow}{HTML}{ffd155}
\definecolor{goodorange}{HTML}{EC4E20}
\definecolor{goodblue}{HTML}{22A8E7}
\definecolor{goodred}{HTML}{FF2384}
\newcommand{\hlpink}[1]{{\sethlcolor{pink}\hl{#1}}}
\newcommand{\hlgreen}[1]{{\sethlcolor{lightgreen}\hl{#1}}}
\newcommand{\hlyellow}[1]{{\sethlcolor{goodyellow}\hl{#1}}}
\newcommand{\dataset}{PANCO\xspace}

\newcommand{\modelname}{HonestBait\xspace}

\newcommand\blfootnote[1]{%
  \begingroup
  \renewcommand\thefootnote{}\footnote{#1}%
  \addtocounter{footnote}{-1}%
  \endgroup
}
\renewcommand{\thefootnote}{\arabic{footnote}}
\title{HonestBait: Forward References for\\Attractive but Faithful Headline Generation}

\author{Chih-Yao Chen$^\ast$ \\ UNC Chapel Hill \\  cychen@cs.unc.edu
        \And  
        Dennis Wu$^\ast$ \\ Northwestern University \\ hibb@u.northwestern.edu \And Lun-Wei Ku \\ Academia Sinica\\lwku@iis.sinica.edu.tw}

\begin{document}
\maketitle
\blfootnote{\hspace*{-4.5pt}$^\ast$ Equal contribution.}
\begin{abstract}
Current methods for generating attractive headlines often learn directly from
data, which bases attractiveness on the number of user clicks and views.
Although clicks or views do reflect user interest, they can fail to reveal how
much interest is raised by the writing style and how much is due to the
event or topic itself. 
Also, such approaches can lead to harmful inventions by over-exaggerating
the content, aggravating the spread of false information. In this work, we
propose \modelname, a novel framework for solving these issues from another
aspect: generating headlines using forward references (FRs), a writing technique
often used for clickbait. 
A self-verification process is included during training to avoid spurious inventions.
We begin with a preliminary user study to understand how FRs
affect user interest, after which we present \dataset\footnote{Data is publicly available at: \url{https://github.com/dinobby/HonestBait}}, an innovative dataset
containing pairs of fake news with verified news for attractive but faithful
news headline generation.
Automatic metrics and human evaluations show that our framework yields more
attractive results (+11.25\% compared to human-written verified news headlines) while maintaining high
veracity, which helps promote real information to fight against fake news. 
\end{abstract}
\label{sec-intro}
\section{Introduction}
Fake news has become a medium by which to spread
misinformation~\cite{oshikawa-etal-2020-survey, acmwebfakenews}.
One common way to fight against fake news is to release verified
news.\footnote{In this work, we define ``verified news'' as news written
specifically to clarify false information; the term ``real news''  is defined
as general news that does not contain misinformation.} However,
as the goal of news verification is to correct misinformation, verified news headlines
are often bland, making it difficult to gain the attention of users, which
works against the need to alleviate the harmful impact of fake news.
Therefore, headlines for verified news articles should be rewritten to be more
intriguing but still faithful, which is expected to pique reader interest in
verified news. Many studies have been
conducted on generating attractive
headlines~\cite{jin-etal-2020-hooks,xu-etal-2019-clickbait}, among which
\textit{clickbait} represents the style that generates the most reads or
clicks.
Despite their success in attracting readers, there are several challenges in
current models. 
First, clickbait datasets for training headline generators with sensational
style transfer are commonly collected based on the amount of views or clicks,
which assumes that headline popularity is always due to 
the writing style~\cite{popularitybased}. However, user reading preferences could also be motivated by
trending topics or major events. For instance, ``\textit{Flights cancelled as
typhoon nears}'' was the most popular news on a day that a typhoon was coming.
Although such headlines get many views and clicks, the writing style itself is
not interesting, and could end up as noise
in the dataset. 
Second, harmful ``hallucinations'' created by headlines exaggerated to be more
sensational could distort the meaning of the original article. This is
especially critical as we do not want our model itself to spread
misinformation. 
However, as such sensational headline generation models often generate
clickbait with more ambiguous words, it increases the difficulty of evaluating faithfulness by aligning
title semantics with the news content. 

In this work, we propose making real news intriguing by learning what fake
news is good at. We seek to learn what makes fake news eye-catching
instead of simply mimicking the titles of fake news. 
Quantity-wise, the many circulating fake news articles serve as learning materials by
which we can learn to generate more attractive headlines; 
style-wise, fake news is deliberately written to attract attention.  
To learn such attractive writing styles, we adopt the forward-reference (FR)
writing technique~\cite{BLOM201587}, which draws from
psychology and journalism, and is frequently used to create attractive headlines.
Specifically, FR creates an information gap between readers and the news content
with the headline, motivating the reader's curiosity~\cite{psychologyofcuriosity} 
to investigate the news content, and hence provoking the desire to click on the
headline.
One example is the headline \textit{``Wanna be an enviable couple? 12 things a
happy couple must do... It's that simple!''}, which drives readers to find out
what those things are. 

Here, to understand the relation between veracity, attractiveness, and FR types
in news headlines, we conducted a preliminary user study to investigate the
attractiveness of fake and real news, and analyzed the FR types
used in headlines in terms of veracity.
Given these results and observations, we propose \modelname, a novel framework
by which to generate attractive but faithful headlines. 
In this framework, we use FR to remove the need to learn directly from the
click-based dataset.
To ensure the faithfulness of the generated headlines, we design a
lexical-bias-robust textual entailment component on the generated headline and
its original content to confirm that the content infers the headline. 
In addition, we propose \dataset, an innovative dataset which consists of pairs
of fake and verified news headlines, their content, and their FR types. We
conduct experiments on \dataset and evaluate the results in terms of both automatic
metrics and human evaluation. 
In sum, the contributions of our work are threefold:
\begin{itemize}
	 \item We conduct a thorough user study to understand the relation between
	 reading preferences and FR types on fake news and verified news. 
	 \item We propose a novel framework for generating attractive but faithful
	 headlines. In human evaluations, \modelname 
	  largely outperforms baselines on attractiveness and faithfulness.
	 \item We propose a new dataset containing pairs of fake and 
	 verified news, including their headlines, content, and FR
	 types in headlines.
\end{itemize}


\label{sec-related}
\section{Related Work}
\subsection{Forward Referencing as a Lure} 
\citet{psychologyofcuriosity} shows how the desire for information motivates
human curiosity. 
Forward-referencing has been defined as a technique for creating curiosity gaps
at a discourse level for use in headlines~\cite{BLOM201587,Yang2011ACI}. 
A similar concept is cataphora, in which information is forwarded as a teaser
at the sentence level~\cite{cataphoric, Halliday76a}. 
\citet{digitalclickbait} investigate how editors rewrite headlines for digital
platforms, and analyze the linguistic features of what makes for an attractive
headline.
\citet{zhang2018question} address attractive headline generation as question
headline generation (QHG), which assumes that interrogative sentences are more
popular. Although this modality is indeed a type of FR, we argue that the
interrogative style may not be suitable for all kinds of headlines, especially
verified news. Hence in our work, we fully consider all kinds of FR which
are commonly used and seen in social media and on digital platforms.
Sample headlines exhibiting FR techniques can be found in
Fig.~\ref{fr-demo} in the appendix.

\subsection{Headline Generation} 
Headline generation can be viewed as a more specific summarization task.
\citet{qi2020prophetnet} propose a Transformer-based, self-supervised n-gram
prediction objective. 
\citet{liu2019finetune} propose BERTSum, a variation of
BERT~\cite{devlin-etal-2019-bert} for extractive summarization.
\citet{see-etal-2017-get} propose an attention-based pointer generator with a
copy mechanism, which has made great progress in summarization. 
Although its ability to copy text from the source context is powerful,
using it directly for verified news often leads to bland titles.
Hence we apply FRs and a sensationalism scorer to produce more satisfying results.
\citet{xu-etal-2019-clickbait} propose auto-tuned reinforcement learning
to generate sensational headlines using a pretrained sensationalism scorer;
the resulting score is used as the reward to enhance the attractiveness.
Although generating attractive headlines has been widely
explored~\cite{popularitybased, jin-etal-2020-hooks}, we focus more on
fidelity to ensure that the semantics of the generated headline are faithful to
the source content to avoid harmful hallucination.

\subsection{Faithful Summarization}
Recent work investigates how to improve the faithfulness of the generated
summary or headline. \citet{Matsumaru2020ImprovingTO} propose 
pretraining a textual entailment scorer to filter out noisy samples
in the dataset, preventing hallucination or unfaithful generation.
\citet{onfaith} analyze the faithfulness of current abstractive summarization
systems, and discover that textual entailment is better correlated
to faithfulness than standard metrics. Based on such work, one
major direction is to evaluate generated summaries in terms of textual entailment
rather than raw metrics such as ROUGE~\cite{lin-2004-rouge} or BLEU~\cite{bleu}.
Accordingly, we propose a faithfulness scorer based on textual entailment to
evaluate how well the generated headlines fit the semantics of the content. 

\section{Preliminary User Study}\label{userstudy}
In this section, we investigate for a given topic which of the fake or real
headlines users are more interested in, and how often forward references are
found in interesting titles. Accordingly, we seek to test the following two hypotheses:

\noindent\textbf{H1}: \textit{Fake news headlines motivate user reading
	 interest more than real news headlines.}
  
\noindent\textbf{H2}: \textit{Forward references are commonly seen/used in headlines which interest users.}

We conducted the user study on both Chinese and English news to determine whether
forward references were used across languages.
For English headlines, we adopted FakeNewsNet~\cite{shu2018fakenewsnet}, which
contains fake and real news headlines about gossip and political news from
GossipCop and PolitiFact. Since the real and fake news in FakeNewsNet are not
paired up, we performed topical clustering to alleviate topical bias.  
For Chinese headlines, we directly leveraged news pairs labeled as
\textit{disagreed} in the WSDM fake news challenge
dataset,\footnote{\url{https://www.kaggle.com/c/fake-news-pair-classification-challenge}}
which contains one fake news headline and its corresponding verified news
headline. 

\begin{figure}[h]
\centering
\subfigure[Chinese headlines]{\label{fig:a}\includegraphics[width=38mm]{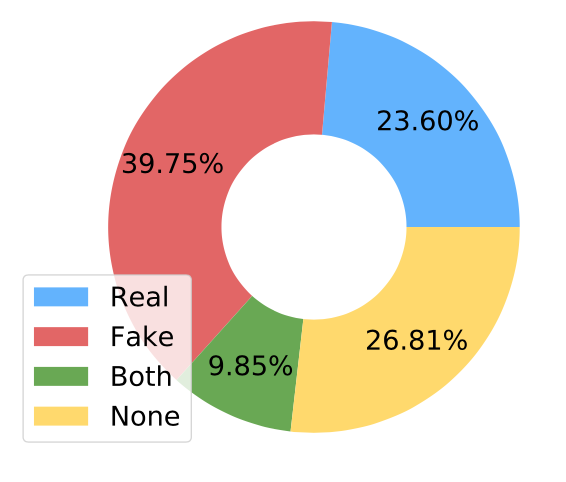}}
\subfigure[English headlines]{\label{fig:b}\includegraphics[width=38mm]{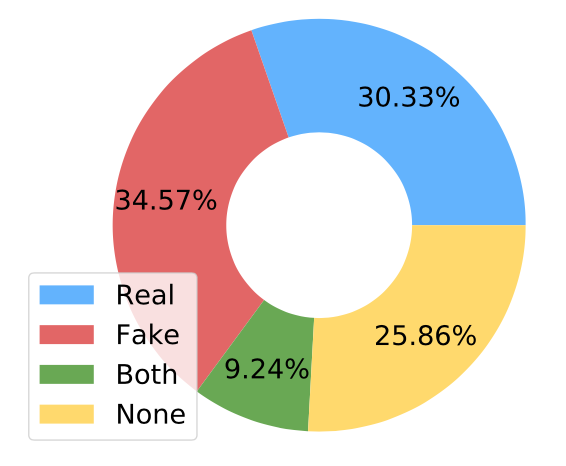}}
\caption{Reading preferences w.r.t.\ real news and fake news including ties.
The sample size is 8,424 / 6,497 for Chinese / English headlines.}
\label{userstudy_dist}
\end{figure}

We conducted the English user study using Amazon Mechanical Turk~\cite{amturk}.
Each pair was labeled by three turkers, whereas each Chinese pair was annotated by
five native speakers we recruited. 
To test H1, annotators chose which headline they wanted to read further, with
four options: \emph{first headline}, \emph{second headline}, \emph{both}, and
\emph{none}. News veracity was not revealed during the study. 
Results show that both Chinese and English readers prefer fake news headlines.
For Chinese headlines, 39.75\% of fake titles were judged to be more
interesting than the real ones, whereas only 23.60\% of real titles won. For
English headlines, the percentages are 34.57\% and 30.33\%, respectively. Note
that in English, we are comparing real news with fake news due to the scarcity
of paired verified and fake news data, whereas in Chinese, we are comparing
verified news with fake news. This could be why the preference for
real and fake news in English is closer than in Chinese. Even so, both Chinese and
English show with statistical significance (p-values far less than 0.05) that
readers prefer fake headlines. We report the complete distribution including
ties, as shown in Fig.~\ref{userstudy_dist}.
This result supports H1: fake news headlines motivate reading interest more
than real news headlines.

To test H2, we randomly sampled 1,000 preferred and rejected headlines,
respectively, from the previous user study, and asked another set of three
annotators to label the FR type. 
Results show that 73.48\% of Chinese and 85.32\% of English preferred headlines
utilizing FR techniques (at least one FR included in the headline), whereas in
rejected headlines, the ratio is 22.35\% / 17.72\%. This further supports
H2: FR is commonly used in interesting headlines. In conclusion, we found that
fake news headlines draw more reader interest, and the use
of FR techniques is a key part of what makes headlines intriguing.

\begin{figure*}
\includegraphics[width=1.0\textwidth]{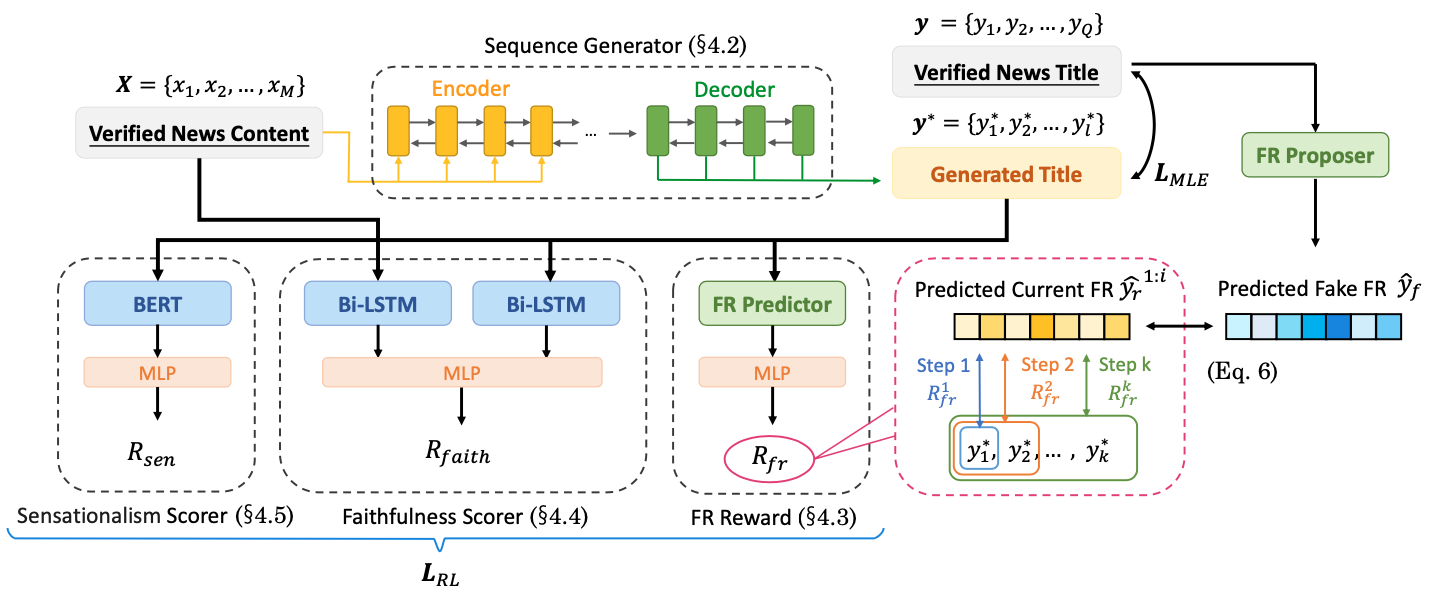}
\centering
\caption{\label{fig_model}Stage 2: overall architecture of \modelname.
Required inputs are underlined.}
\centering
\end{figure*}
\section{Methodology}
\label{sec-model}
Having motivated the use of FR, we propose \modelname, a novel framework which
incorporates FR techniques and veracity verification. \modelname consists of
two stages. In the first stage, we pretrain an FR predictor
and an FR proposer~(\S~\ref{subsec_fr_scorer}). Both of them take verified news titles as input. 
The FR predictor is trained to predict which FRs a verified headline contains;
hence the gold label is the FRs of the current input verified headline. The FR
proposer, in turn, learns to predict which combination of FRs the corresponding fake
news exhibits; the gold label is the FRs of the corresponding fake
headline of the current input verified headline. The main concept in stage~1 is
learning FRs from fake news to provide the direction best suited
to rewriting a monotonic verified headline into an interesting headline. 

When the FR predictor and FR proposer are ready, we proceed to stage~2 to
generate attractive but faithful headlines. 
Figure~\ref{fig_model} depicts the overall architecture of the second stage. 
The input during stage~2 consists only of verified news headlines and their
content for learning headline generation, where the developed FR predictor
and FR proposer together provide rewards to the learning model. First, we use a
sequence generator~(\S~\ref{subsec_seq_gen}) to generate headlines from the
input verified news content, and utilize the FR proposer to predict which
combination of FR types is best suited to rewriting the input verified news
headline. 
During each decoding step, we use the FR predictor to predict which FR types 
the currently generated headline contains, and we align the prediction
from the FR proposer and the FR predictor to transform the
original boring verified headlines into exciting ones. This is achieved by
computing the FR type reward~(\S~\ref{subsec_fr_reward}). 
After decoding, we make use of a faithfulness
scorer~(\S~\ref{subsec_faithfulness_scorer}) and a sensationalism
scorer~(\S~\ref{subsec_sensation_scorer}) to compute the faithfulness and
sensationalism rewards by which to evaluate the generated headline; all
three rewards are then combined to make the generated results attractive but
faithful.
During inference, given verified news headlines and their content, \modelname
then generates attractive but faithful headlines using the above-mentioned
components. 
Below we describe each major component in detail.
\subsection{FR Predictor \& FR Proposer}
\label{subsec_fr_scorer}
To mimic different FR types on datasets without FR type
labels, we pretrain two multi-label classifiers: 
(1)~A FR predictor, which predicts which FR type the generated headline contains;
this is pretrained by taking verified news headlines as input and classifying
which FR type these headlines exhibit. 
(2)~A FR proposer, which learns what specific combination of FRs is best suited to
rewriting a given verified title.
This is trained by taking the verified headline as input and predicting the FR
type of the corresponding fake news.
Note that this setting is achievable because we have paired news data with both
real and fake FR labels (see preview sample in Fig.~\ref{PANCO_sample} in the
appendix).

We implement these FR classifiers with a BERT-based encoder. Given a verified
news headline, we obtain a sentence-level representation $h_p$ with the hidden
state of the [CLS] token. 
The FR type $\hat{y_{\mathit{fr}}}$ is predicted by a MLP classifier following
a sigmoid function and a softmax operation: 
$\label{eq_fr}
\hat{y_{\mathit{fr}}} = \mathrm{softmax}(\sigma(W_{p}h_{p}+b_{p})),$
where $\hat{y_{\mathit{fr}}} \in \{0,1\}^{l}$, $l$ is the number of the FR type, and $W_p$,
$b_{p}$ are trainable parameters. 
We pretrain these models using binary cross entropy loss, yielding a 0.91
micro-F1 score for the FR predictor and 0.65 for the FR proposer on a pretraining test
set. Below we denote the FR predictor's prediction as $\hat{y}_{r}$ and the FR proposer's prediction
as $\hat{y}_{f}$. 

Predicting the fake version of FR types from the verified news headline is more
challenging, as the performance of the FR proposer is lower than of the FR predictor (0.65
vs.\ 0.91).
In practice, we could directly use the FR label of the fake news acquired from
our user study to replace $\hat{y}_{f}$, and view this setting as an upper
bound for the FR proposer accuracy to calculate $\mathcal{R}_{\mathit{fr}}$.
However, when we are not provided with the FR labels of fake titles, we do not
know which FR technique(s) should be applied to rewrite the given
verified news headline. Hence, the FR proposer can be used as an auxiliary tool to
help decide which FR type to use; this is especially useful when the dataset
contains no FR-type labels.
After pre-training the FR predictor and proposer, we proceed to the second
stage. 

\subsection{Sequence Generator}
\label{subsec_seq_gen}
In the second stage, we adopt a pointer network~\cite{see-etal-2017-get} as the
sequence generator because of its ability to copy words from the source text.
Given verified news content with $M$~tokens $ X=\{x_1, x_2, \ldots, x_M \}$
and its corresponding real headline consisting of $Q$~tokens $ y=\{ y_1, y_2,
\ldots, y_Q \}$, the encoder encodes each token with a bidirectional
LSTM~\cite{1997lstm}.
We adopt Chinese word-level embeddings pretrained on the Weibo corpus~\cite{2018chemb}. The final distribution is
combined with the probability computed by the copy mechanism, making words from source content available for generation. For the objective we use the negative log likelihood as
\begin{equation}
    \mathcal{L}_{\mathit{MLE}} = - \frac{1}{T} \sum_i^{T} \log P_{\mathit{final}}(y_i).
\end{equation}

\subsection{Forward Reference Reward}
\label{subsec_fr_reward}
For each decoding time step, we calculate the FR reward once:
tokens generated up to the current time step $y^{*}_{1:t}$ are sent to the FR
predictor to derive $\hat{y}_{r}^{1:t}$ and calculate how well the generated
text fits the FR prediction using the FR Proposer $\hat{y}_{f}$. 
  After $T$~steps of decoding, and after the headline is generated,   
we calculate the average FR reward as
\begin{equation}
    \mathcal{R}_{\mathit{fr}} = \frac{1}{T}\sum_i^T(1 - \mathcal{D}(\hat{y}_{f}, \hat{y}_{r}^{1:i})),
\end{equation}
where $\mathcal{D}$ denotes a distance function---in our case the mean squared
error---and $\mathcal{R}_{\mathit{fr}} \in [0, 1]$ is the average FR reward. 
Here $\hat{y}_{r}$ is the FR types
exhibited by the current generated headline, which should align
with the prediction from the FR proposer $\hat{y}_{f}$, which is pretrained to learn
which specific combination of FRs are best suited to rewrite the given title. The
closer they get, the higher $\mathcal{R}_{\mathit{fr}}$ is. 

\subsection{Faithfulness Scorer}
\label{subsec_faithfulness_scorer}
Inspired by research which shows that textual entailment correlates better with
faithfulness than raw metrics~\cite{falke-etal-2019-ranking}, we use a
pretrained faithfulness scorer to evaluate whether the generated headline
distorts or contradicts the corresponding content. 
When pretraining, we use a verified news headline and its content as a
positive example, and use a fake news headline with the corresponding real news
content as a negative example. 
We pretrain this as a natural language inference (NLI) task (classifying
entailment and contradiction). 
The headline and content sentence embeddings of are denoted as $x_f$ and $w_f$.
We apply a popular method to encode sentences for the NLI model~\cite{infersent}:
\begin{equation}
    h = [x_f; w_f; x_f - w_f; x_f \odot w_f],
    \label{infersent}
\end{equation}
where ``$;$'' denotes concatenation, and ``$\odot$'' denotes the element-wise product. The faithfulness scorer achieves an accuracy of 0.83 on the testing set.

\subsection{Sensationalism Scorer}
\label{subsec_sensation_scorer}
Apart from the FR type reward, we make use of another BERT-based binary
classifier to obtain the sensationalism score, since there are headlines that
are still interesting without the use of FRs (around 27\% according to our
collected data). 
We first manually reviewed 100 news for each categories in seven different new
sources, selected the news categories that were consistently sensational (more
than two-thirds of the articles in such a category were sensational, e.g., fashion, gossip,
headlines) and collected the news headlines along with the content in these
categories. We reviewed 5,000 headlines in total and collected 50,000
sensational news headlines.
For non-sensational headlines, we utilized a pointer generator to obtain a
summary headline, and treated this as a non-sensational title since summarization
models retain only the semantics of the content. In this way we ensured a 50/50
split for sensational and non-sensational headlines for training.
We trained the sensation scorer using binary cross entropy along with a softmax
layer to produce a sensationalism score $\in [0, 1]$:
$\mathcal{R}_{\mathit{sen}} = \sigma( W_{s} h_s + b_s ), $
where ${h_s}$ is the aggregated representation of the [CLS] token produced by
BERT, and $W_s$ and $b_s$ are learnable weights.
The accuracy on the test set is 0.86, indicating its ability to discriminate
sensational headlines.

\subsection{Hybrid Training}
We adopted reinforcement learning (RL)~\cite{rl1992} to train our model
with the weighted sum of $\mathcal{R}_{fr}$, $\mathcal{R}_{faith}$ and $\mathcal{R}_{sen}$ as the reward~$\mathcal{R}$. Following
\citet{xu-etal-2019-clickbait, ranzato2015sequence}, we used the baseline reward
$\hat{\mathcal{R}_t}$ to reduce variance, where $\hat{\mathcal{R}_t}$ is the
mean reward estimated by a linear layer for each time step~$t$ during training.
The final reward and the objective are 
\begin{equation}
\begin{gathered}
    \mathcal{R} = \mathcal{R}_{\mathit{fr}} + \alpha\mathcal{R}_{\mathit{faith}} + (1-\alpha)\mathcal{R}_{\mathit{sen}} \\[4pt]
    \mathcal{L}_{\mathit{RL}} = -\frac{1}{T} \sum_i^T (\mathcal{R}-\hat{\mathcal{R}_t})\log P_{\mathit{final}}(y_t). \\[4pt]
\end{gathered}
\end{equation}
Similar to \citet{xu-etal-2019-clickbait}, we computed the final loss as the
combination of $\mathcal{L}_{\mathit{MLE}}$ and $\mathcal{L}_{\mathit{RL}}$:
\begin{equation}
\begin{aligned}
    \mathcal{L} = \lambda \mathcal{L}_{\mathit{MLE}} + (1- \lambda)\mathcal{L}_{\mathit{RL}},
\end{aligned}
\end{equation}    
where $\alpha$ and $\lambda$ $\in [0,1]$ are hyperparameters that balance
the weight of each component; the composite design here ensures that
we produce headlines that satisfy all objectives. In sum, we use the FR reward
to estimate whether the generated headline matches the FR type of its fake
version, the faithfulness scorer to evaluate the textual entailment between the
generated headline and the verified news content, and the sensationalism scorer
to measure the sensationalism of the generated headline. 

\label{sec-experiment}
\section{Experiment}
In this section, we describe experiments conducted to evaluate \modelname.
We first describe the experimental dataset and then describe the result of human evaluation,
automatic metrics, a case study, and hyperparameter analyses 
to further demonstrate the superiority of the proposed model. 

\subsection{PANCO Dataset}

\label{sec-dataset}
We collected \textbf{Pa}ired \textbf{N}ews with \textbf{Co}ntent (PANCO), 
a subset of a fake news classification competition held by WSDM.
The competition involved a textual entailment task in which two news headlines
were given as input: the task was to predict the relationship between the
headlines. Each sample in the original dataset included a fake news headline
and a headline that was either \textit{agreed} (two fake stories describing the
same event), \textit{unrelated} (two stories describing different events), or
\textit{disagreed} (two stories describing the same event, one of which was
fake and the other was verified).
We selected the \textit{disagreed} pairs that contain a fake headline and its
corresponding verified news headline, and augmented the dataset in the
following way:
(1)~We used each title as a query which we submitted to Google Search to
determine the source of each news story, and crawled the news content from
sources which matched the title. 
(2)~Five annotators labeled the FR type of each headline; the final label was
decided by majority vote. 

The proposed dataset consists of a total of 7,930 paired samples containing a fake news
headline and the corresponding verified news headline along with their content
and FR type. To better understand the dataset, we provide a preview sample in
Fig.~\ref{PANCO_sample} in the appendix. The main novelty of \dataset is the
collection of pairs (describing the same event) of fake and verified news
with headlines and their content. In addition, we provide the FR type label for
both verified and fake news as additional text features for further study. We provide a previewing sample from PANCO in Table~\ref{PANCO_sample}.

\begin{figure*}
    \centering
    \includegraphics[width=0.8\textwidth]{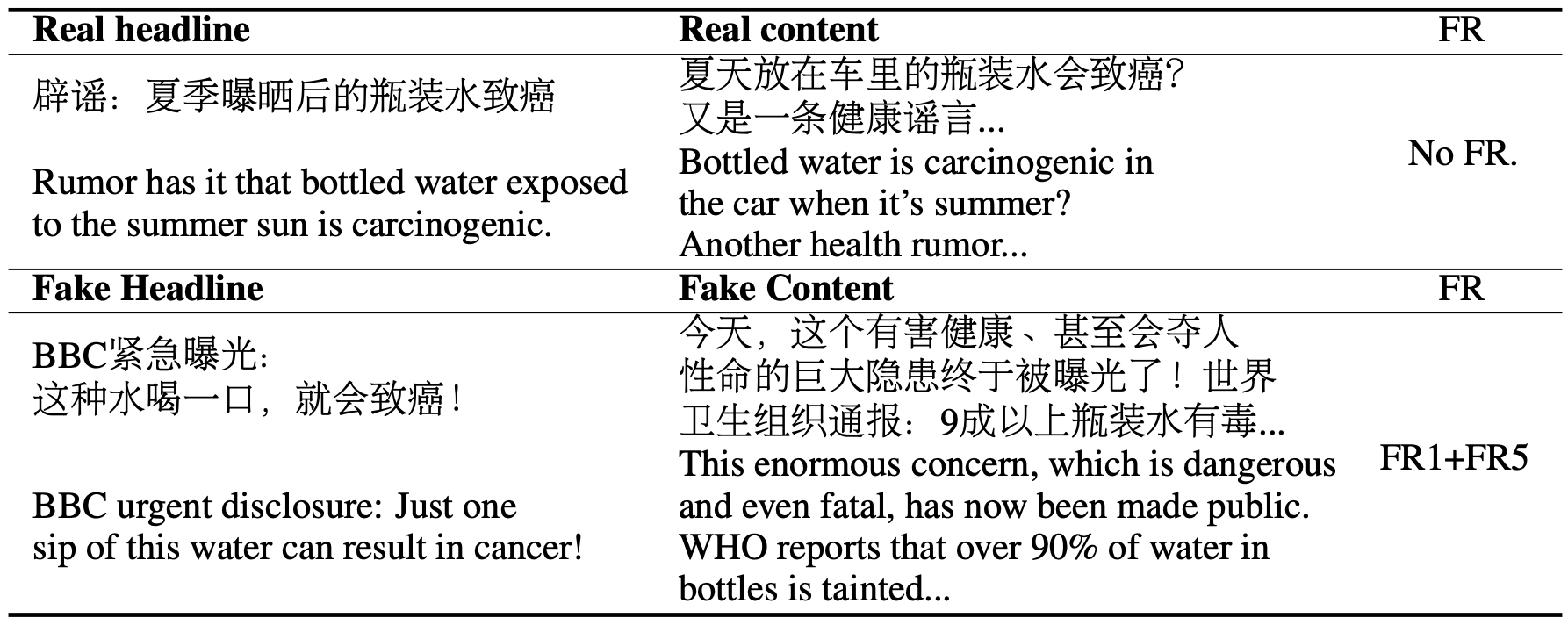}
    \caption{A previewing sample from PANCO dataset,
comprised of paired real and fake news headlines, the news content,
and FR type labels for both real and fake headlines.}
    \label{PANCO_sample}
\end{figure*}



\begin{table*}[h]
\center
\small
\begin{tabular}{ll}
\bottomrule
\\[-1em]
Reference & \begin{tabular}[c]{@{}l@{}}Verified news headline: The truth of using one drop of blood to test cancer. \\ 
Fake news headline: \hspace{9pt} Testing cancer using only one drop of blood! This is amazing. \end{tabular}\\ \\[-1em]
\hline
\\[-1em]
News content       & \begin{tabular}[c]{@{}l@{}}``The woman version of \hlyellow{Jobs}'' Elizabeth Holmes became popular by proposing a revolutionary\\technique: using a single drop of blood to test cancer. But not for long: her lies were revealed,\\and she fell from favor. \hlgreen{An expert said} that liquid biopsies in clinics cannot yet be consisted\\the gold standard, and cannot completely replace tissue biopsy.\end{tabular} \\ \\[-1em] \hline \hline \\[-1em]
Ptr-G              & Rumors! \hlyellow{Jobs}, really tells you the truth! \\ \\[-1em] \hline \\[-1em]
Clickbait          & ``Rumors'' \hlyellow{Jobs} can heal the reagent box? Here’s the truth! \\ \\[-1em] \hline \\[-1em]
Clickbait+ROUGE    & Rumors! Can \hlyellow{Jobs} make people test for cancer? \\ \\[-1em] \hline \\[-1em]
BERTSum            & A drop of blood can detect cancer?     \\ \\[-1em] \hline \\[-1em]
T5                 & A drop of blood can detect cancer is a rumor, how can we do to \hlpink{prevent cancer}?                     \\ \\[-1em] \hline \\[-1em]
ProphetNet         & Clarification: Blood test cannot determine cancer.                                           \\ \\[-1em] \hline \\[-1em]
HonestBait         & A drop of blood can detect cancer? \hlgreen{Experts clarified}: it’s a scam!
\\ \\[-1em] \toprule
\end{tabular}
\caption{\label{casestudy}
Generated examples from different models. For brevity, we show part
of the article and translated result.
}
\end{table*}

\subsection{Baseline and Settings}
We compared the proposed model with the following strong baseline for headline generation. 
\noindent\textbf{Ptr-G} for pointer generator
	 network~\cite{see-etal-2017-get}, an LSTM-based model with
	 attention and a copy mechanism.
\noindent\textbf{Clickbait}~\cite{xu-etal-2019-clickbait}, which uses a CNN-based
	 sensationalism scorer to automatically balance MLE and reward loss, and also used as a reward to generate more sensational headlines.
\noindent\textbf{ROUGE}, which uses the same architecture as Clickbait but with the
	 ROUGE score as a reward.
\noindent\textbf{BERTSum}~\cite{liu2019finetune}, which utilizes BERT's architecture
	 to encode source text and perform extractive summarization.
\noindent\textbf{T5}~\cite{2020t5}, a large Transformer-based model;
	 we utilize T5 with PEGASUS~\cite{zhang2020pegasus} pretraining to
	 strengthen the baseline.
\noindent\textbf{ProphetNet}~\cite{qi2020prophetnet, qi2021prophetnet}, a
	 Transformer-based model that utilizes future n-gram prediction as a
	 self-supervision. 

For human evaluation and the case study, we also include Gold, which represents
human-written verified headlines as a strong baseline. Experimental settings
are detailed as follows.
We first pretrained all baselines on the LCSTS
dataset~\cite{hu-etal-2015-lcsts} with 480,000 steps. LCSTS is a large-scale
Chinese summarization dataset containing 2,400,591 samples with paired short
text and summaries. We used the pretrained weights to fine-tune all baselines
on the \dataset training set for another 20,000 steps. We saved the checkpoints
for all baselines every 2,000 steps, and compared them by selecting the
best one on the validation set. The hyperparameters of \modelname were also based on
the validation results: $\lambda=0.2$ and $\alpha=0.4$. 

\begin{table}[h]
\centering
\small
\begin{tabular}{lccccc}
\bottomrule
Model            & $R_1$   & $R_2$   & $R_L$   & BS & FR \\ \hline
Ptr-G                     & 41.86          & 28.18          & 37.30          & 69.61               & 55.32                                  \\
Clickbait                 & 41.02          & 28.03          & 36.64          & 69.52               & 69.11                                  \\
ROUGE           & 43.75          & 27.65          & 35.65          & 71.56               & 58.91                                  \\
ProphetNet                & \textbf{46.82} & 30.40          & 38.89          & \textbf{73.57}      & 49.77                                  \\
BertSum                   & 28.09          & 16.15          & 18.86          & 63.22               & 16.83                                  \\
T5                        & 44.27          & 28.55          & 38.66          & 72.73               & 59.96                                  \\ \hline
\modelname & 43.76          & \textbf{31.45} & \textbf{40.42} & 72.61               & \textbf{80.42}                                  \\ \toprule
\end{tabular}
\caption{\label{auto}
Automatic metrics of proposed model against baselines. $R_n$ is the
n-gram ROUGE score, $R_L$ is the ROUGE-L score, BS is the BERT score,
and FR is the ratio of the generated headlines using FR.
}
\end{table}

\begin{table}[!h]
\centering
\small
\begin{tabular}{lccc}
\bottomrule
Model  & ATRC & FAITH & FLCY \\ \hline
Ptr-G           & -29.50\%        & -17.83\%         & -19.80\%        \\
Clickbait       & ~-6.00\%        & -22.33\%         & ~-9.25\%         \\
ROUGE & -17.50\%        & -17.25\%         & -24.66\%        \\
BertSum         & -30.50\%        & -21.99\%         & ~-9.70\%         \\
T5              & -12.50\%        & -10.25\%         & ~-1.25\%         \\
ProphetNet      & ~-5.60\%        & ~-5.50\%         & ~~4.33\%          \\
Gold (human)            & -11.25\%        & ~~1.00\%         & ~~8.34\%         \\ 
\hline
HonestBait      & -             & -              & -             \\ \toprule
\end{tabular}
\caption{\label{human-eval} Pairwise comparison in terms of
attractiveness (ATRC), faithfulness (FAITH), and fluency (FLCY), shown as
percentages. The larger the negative value, the more \modelname outperforms.}
\end{table}

\subsection{Human Evaluation}
\label{sec5_4}
We first conducted a human evaluation to evaluate the attractiveness,
faithfulness, and fluency of the generated headlines. 
We randomly selected 100~samples from the \dataset test data, and asked five native speakers to select
headlines in response to the following questions: 
(1)~which headline makes you want to read further? 
(2)~which headline is more faithful to the content?
(3)~which headline is more fluent?

The workers were given two generated titles and the story content, and were asked
to select \textit{first title}, \textit{second title}, or \textit{tie} in
response to the questions. 
Table~\ref{human-eval} reports the pairwise comparison results as percentages.
Each number in the table is the competing model compared to the proposed
\modelname, following \citet{zhao-etal-2020-bridging}. For
example, the output of Ptr-G is 12.50\%/45.50\%/42.00\%
better/same/worse than \modelname in terms of attractiveness, resulting in
$12.50\%-42.00\% = -29.50\%$ in the table. Results show that for both
attractiveness and faithfulness, \modelname
outperforms all baselines by a large margin. We believe this is due to
the use of forward referencing and the faithfulness check. 
Compared to the pure click-driven attractiveness-optimized 
Clickbait~\cite{xu-etal-2019-clickbait}, \modelname outperforms by directly learning 
writing skills to avoid other impact factors of attractiveness.
In addition, boosting only attractiveness makes Clickbait relatively
unfaithful (-22.33\%). 
In terms of fluency, only ProphetNet and human-written headlines outperform our
model. As we did nothing specifically to improve fluency such as ProphetNet's n-stream
attention, this result indicates that \modelname maintains
reasonable fluency while increasing attractiveness and faithfulness. 
Note that compared to human-generated real headlines, \modelname generates
more attractive headlines (+11.25\%) with only a modest drop in faithfulness
(-1.00\%). 
These results show the effectiveness of \modelname for rewriting real news
headlines to promote stories, as it maintains high faithfulness while being
more attractive.
 


\subsection{Automatic Metrics}
We used three automatic metrics for evaluation:
\mbox{ROUGE-n}~\cite{lin-2004-rouge}, ROUGE-L, and the BERT
score~\cite{bert-score}. Although in general, automatic metrics are shown to be not
reliable for text
generation~\cite{sulem-etal-2018-bleu,callison-burch-etal-2006-evaluating,schluter-2017-limits,wang-etal-2018-metrics},
we still provide them here for reference. The results in Table~\ref{auto} still
show the good abstractive ability of \modelname with the highest 40.42 $R_L$
score. Among the baselines, ProphetNet is the strongest, with the highest $R_1$
and BERT scores, perhaps due to its n-stream self-attention mechanism. 
However, the extractive summarization model BERTSum performs worst
here, as extracting a sentence from the article as its headline is not a common
practice in general. 
In the last column of Table~\ref{auto}, we further use the FR predictor to
detect which FR technique(s) the generated headlines are using, and report the
percentage of generated headlines that use FR. The result shows that 80.42\% of
the headlines generated by \modelname exploit FRs to make headlines more
attractive, which is the highest among all models, indicating that \modelname
indeed learns to utilize FR techniques.

\subsection{Ablation Study}
To further investigate our framework, we conducted an ablation study. 
We compared each setting with the full framework using the evaluation protocol
from \S~\ref{sec5_4} by pairwise comparison, along with the automatic metrics
for completeness. 
The results are shown in Table~\ref{ablation}. 
Clearly, there is a significant drop in attractiveness when we remove the
sensation scorer (-19.50\%) or FR type reward (-16.00\%), which indicates that
even with the sensation scorer, attractiveness still decreases without the help
of the FR reward (see setting \emph{w/o FR}). That is, the FR reward indeed
helps the model to learn attractive writing styles. 
In addition, removing the faithfulness scorer results in the largest decrease in
faithfulness (-11.50\%). This also shows that our faithfulness
scorer prevents deviations in the generated headline. Interestingly, removing
the sensation scorer increases the ROUGE score, perhaps because the
sensation scorer helps to generate more diverse and interesting headlines, and thus
can harm metrics which are based on word-level overlap. We also observe that
removing the faithfulness scorer reduces the ROUGE score, which shows that
the faithfulness scorer helps to produce headlines with more fidelity, and
thus increases the word-level overlap between the generated headlines and the 
ground-truth.
Note that as automatic metrics are still not the most important indicator of
generation quality, thus we still keep sensation scorer for its improvements in terms of attractiveness and
fluency even if removing the it leads to a higher
ROUGE score.

\begin{table}[!h]
\centering
\small
\begin{tabular}{ cccccc }
\bottomrule
 & ATRC & FAITH & FLCY & $R_2$ & \\
\hline
W/o sen   & -19.50\% &   -4.00\%  & -9.75\% & \textbf{32.01} \\
W/o faith &  -4.00\% &	-11.50\%  & -6.75\% & 28.81 \\
W/o FR    & -16.00\% &  - 5.50\%  & -6.25\% & 30.92 \\
Full      &   -      &     -      &    -    & 31.45 \\
\toprule
\end{tabular}
\caption{\label{ablation}
Ablation study result. 
}
\end{table}

\subsection{Case Study}
Table~\ref{casestudy} shows an example illustrating headlines generated by
different models. 
Results show that Ptr-G, Clickbait, and ROUGE extract the name ``Jobs'' from
the article (highlighted in yellow), which is a powerful ability of the copy
mechanism to alleviate the generation of unknown tokens.
However, in terms of being headlines, these texts are less satisfying in that
they are not understandable. BERTSum and T5 make mistakes by generating open
questions without answering them, which could motivate user interest but is not
faithful enough for verified news headlines. Even more, T5 focuses on the wrong
point borrowed from other articles as this article is not about cancer
prevention, 
which could be harmful (highlighted in pink). In contrast, \modelname
generates interrogative sentences to attract readers, but with an explicit
clarification of the fake information, and is aligned to the content
(highlighted in green).
\section{Conclusion}
\label{sec-conclusion}

We present \modelname, a novel framework for generating faithful
but interesting headlines from a new aspect: forward references. Moreover, we
construct \dataset, a novel dataset that includes the title and content of
pairs of fake and verified news, along with their forward reference types for
further research. Our user study shows that verified news headlines are
relatively boring, and forward references are used in most headlines liked by readers.
Experimental results show that \modelname outperforms all
baselines in both automatic and human evaluations, which demonstrates its
effectiveness in generating attractive but faithful headlines. We expect
\modelname to help rewrite monotonous real news headlines to increase their
exposure rate to help combat fake news.



\section*{Limitations}
Although \modelname shows promising results for generating attractive but faithful headlines, there are still
some limitations: (1)~\modelname is a monolingual model that only
supports Chinese. It requires three pre-trained scorers. Also, as the FR labels are
specifically difficult to obtain, it is not easy to implement in
other languages. (2)~Running the whole framework with a batch size of 16
takes around 22\,GB GPU memory, mostly because we must load all pre-trained
models into the GPU. This can be alleviated by using a distilled pre-trained model. (3)~On average, \modelname generates more faithful
headlines than other baselines, but it still occasionally produces false
information or unwanted results. This work is only for academic purposes and is
not ready for production.

\section*{Ethics Statement}
Given that our dataset is in Chinese and requires a profound understanding of forward referencing for annotation and evaluation, we carefully selected annotators from our lab who specialize in NLP-related research and possess knowledge in linguistics. To ensure fairness, we provided all annotators with a payment of \$6.66 per hour, which is 10\% higher than the minimum hourly wage requirement in Taiwan.

During the data annotation process, we introduced the concept of forward referencing to the annotators, along with relevant examples. Only annotators who achieved an accuracy rate of over 80\% were eligible to perform the actual annotation task. It's important to note that we solely asked annotators to label the "type of forward reference," which is well-defined, and not to assess the accuracy or truthfulness of the news articles. With five annotators who successfully passed the pretest, combined with the relatively objective nature of labeling forward reference types, we believe any potential bias during the data annotation process is minimal.

For the evaluation phase, an additional five annotators were tasked with determining the preferable title among two options, based on attractiveness, faithfulness, and fluency. These annotators are different from those who labeled the data to ensure a blind test. Although this task involves a greater level of subjectivity, we provided average statistics based on the assessments of the five annotators. Additionally, we maintained a blind test by recruiting separate evaluators and randomly shuffling the order of the two titles for each trial. This evaluation protocol aligns with standard practices employed in the research community, and we believe it effectively minimizes potential biases.

It is also important to note that we are not really \emph{learning to mimic fake news},
by taking fake news headlines as the ground truth reference. Instead, we seek to
learn the \emph{writing techniques} that are often used in fake news to attract
readers. As we are aware of the risk of producing misinformation, we want to
again highlight the importance of the faithfulness check. \modelname was 
designed only to assist journalists as a reference to write 
faithful headlines that users prefer for verified news. Even if we propose using a faithfulness
scorer to increase fidelity, its nature, similar to attractive headline
generation systems, still exhibits the risk that \modelname could be used by
malicious users to generate sensational headlines for fake news. 
Additionally, \modelname may misjudge offensive or unethical headlines to be
a headline that users would prefer. 
Our goal is 
  to fight fire with fire by leveraging   
fake news as learning material to
fight against misinformation, by encouraging users to read verified news. We call on users not to abuse \modelname to produce false information.
\section*{Acknowledgement}
This work is supported by the National Science and Technology Council of Taiwan under grants 111-2221-E-001-021 and 111-2634-F-002-022.
\bibliography{custom}
\bibliographystyle{acl_natbib}

\appendix
\section*{Appendix}
\begin{figure*}[h!]
    \centering
    \includegraphics[width=0.75\textwidth]{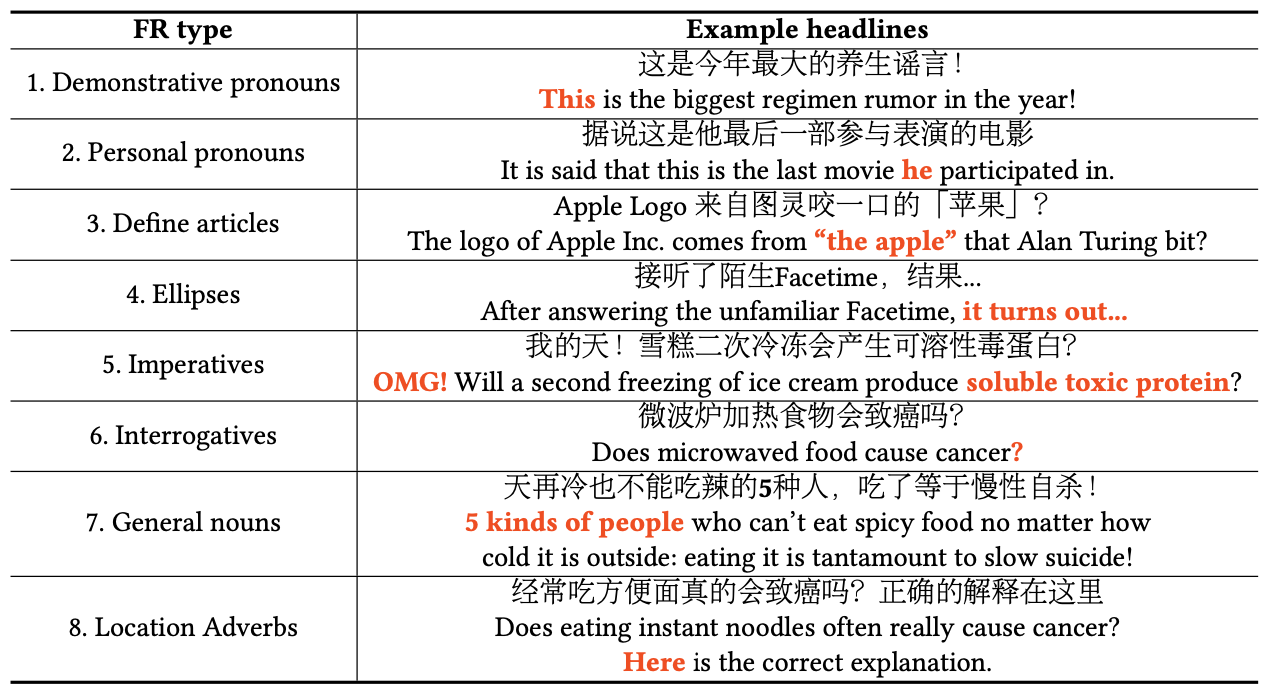}
    \caption{Examples of different types of forward references. Words highlighted in orange are the main characteristics.}
    \label{fr-demo}
\end{figure*}

\section{Debiasing Faithfulness Scorer}
We observe that lexical bias affects our faithfulness scorer. In particular, many verified headlines entailed by their contents
contain the words ``verification'' or ``rumor'' which results in a shortcut
model, i.e., the NLI model tends to classify samples as entailment based simply
on the existence of these certain words. 
In addition, the word-overlap bias (WOB)---classifying hypothesis and
premise as entailment because of high
word overlap~\cite{naik-etal-2018-stress}---also harms our entailment task to
ensure faithfulness. 
Thus, we follow~\citet{debias} in adopting a model-level debiasing module for 
pretraining entailment. A bag-of-words
(BoW) sub-model is deployed to capture superficial features, since it has the
least reasoning ability, and is more likely to use shortcuts to make
predictions. 
The main NLI model, in turn, consists of two bi-LSTM networks
that are capable of reasoning over deeper semantics. 
During training, HEX projection~\cite{wang2018learning} is used to screen out
superficial features by making the hidden state of the main NLI model and the
BoW sub-model orthogonal, forcing the main classifier to focus on deeper
semantic features.

\section{Analysis of Hyperparameters}\label{lam-anal}
\begin{table}[h]
\small
\begin{tabular}{c|c}
\bottomrule 
\\[-1em]
\textbf{$\lambda$} & \textbf{Generated sample headlines}             \\ \\[-1em] \hline
\\[-1em]
0.2                & \begin{tabular}[c]{@{}c@{}}Apples from Linyi county are unsalable? Linyi \\
county government clarifies: \textcolor{goodorange}{\textbf{over-exaggerating}}.
\end{tabular} \\ \\[-1em] \hline \\[-1em]
0.6                & \begin{tabular}[c]{@{}c@{}}Are apples from the county of Linyi unsellable? \\  \textcolor{goodorange}{\textbf{This story is a rumor!}}\end{tabular} \\ \\[-1em] \hline \\[-1em]
1.0                & \begin{tabular}[c]{@{}c@{}}Apples from Linyi county are unsalable? \\ e-commerce's customer service: \\ the merchant may violate portrait rights. \end{tabular} \\[-1em] \\  \toprule    
\end{tabular}
\caption{Headlines generated with different $\lambda$. Orange words are more sensational expressions.}
\label{tab:examp_lamb}
\end{table}

Here we provide a qualitative analysis to examine the sensitivity of $\lambda$
and $\alpha$; recall that $\lambda$ balances MLE loss and RL loss, and $\alpha$
influences the sensationalism. In a sense, a higher $\lambda$ leads to robust yet
boring generation, as a higher $\lambda$ relies more on MLE, and MLE loss is
calculated according to the gold title. Table~\ref{tab:examp_lamb} summarizes
title generation with different $\lambda$.
Note that $\lambda=0.0$ is ignored, as it completely relies on RL loss, which
often leads to broken generation results and is not practical in general. When
$\lambda=1.0$, the model relies completely on MLE loss and is identical to
using only Ptr-G. A smaller $\lambda$ creates more diversity, and $\lambda=0.2$
balances diversity, attractiveness, and fluency. Also, in $\lambda=0.2$ and
$\lambda=0.6$, more sensational or eye-catching words are used (highlighted in
orange in Table~\ref{tab:examp_lamb}), whereas $\lambda=1.0$ shows a 
plain, ordinary tone. When $\lambda=1.0$, the generated results are unrelated
and unintelligible, which also shows that our faithfulness scorer helps align
headline to content, since there is no faithfulness reward when $\lambda=1.0$.

We also conducted an analysis of how different values of $\alpha$ affect the
generated headline. In Table~\ref{tab:examp}, a lower $\alpha$ indicates that a greater
emphasis is put on sensationalism.
A higher $\alpha$ yields a relatively simple and monotonous sentence structure.
In Table~\ref{tab:examp}, $\alpha=1.0$ predominantly generates affirmative
sentences including ``can'', ``is'' or ``will'', which are highlighted in red.
On the other hand, a less dominant $\alpha$ provides more flexibility with
respect to the sentence structure and adds diversity. 
When the reward is completely provided by the sensation scorer and the FR type
reward ($\alpha = 0.0$), it seems that the model generates headlines from a
different aspect and focuses on different keywords (highlighted in blue).
However, such diversity comes at the risk of spurious invention. When $\alpha =
0.0$, the generated result is similar to the tone of fake news, which creates a
clickbait without specifying the facts.
When $\alpha = 0.4$, the generated headlines maintain high veracity while
improving attractiveness. Accordingly, we use $\lambda=0.2$ and $\alpha=0.4$ as
our default setting.

\begin{table}[h]
\centering
\small
\begin{tabular}{c|c}
\bottomrule
\\[-1em]
\textbf{$\alpha$} & \textbf{Generated sample headlines}         \\ \\[-1em] \hline \\[-1em]
0.0               & \begin{tabular}[c]{@{}c@{}}\textcolor{goodblue}{\textbf{How much harm}} will new clothes do to our body?!\end{tabular}                            \\ \\[-1em] \hline \\[-1em]
0.4               & \begin{tabular}[c]{@{}c@{}} Rumor has it that formaldehyde\\in new clothes causes cancer.\end{tabular}                     \\ \\[-1em] \hline \\[-1em]
0.8               & \begin{tabular}[c]{@{}c@{}}Formaldehyde \textcolor{goodred}{\textbf{can}} cause cancer. \end{tabular}                      \\ \\[-1em] \hline \\[-1em]
1.0               & \begin{tabular}[c]{@{}c@{}}Formaldehyde \textcolor{goodred}{\textbf{is}} a carcinogen.\end{tabular} \\[-1em] \\ \toprule
\end{tabular}
\caption{Generated headlines with different $\alpha$. Blue words are more
diversified expressions, and red words are monotonic affirmatives.}
\label{tab:examp}
\end{table}

\end{document}